\documentclass{article}
\usepackage{ijcai20}

\usepackage{times}
\usepackage{soul}
\usepackage{url}
\usepackage{hyperref}
\usepackage{xcolor}
\definecolor{darkblue}{rgb}{0, 0, 0.4}
\hypersetup{colorlinks=true,citecolor=darkblue, linkcolor=darkblue, urlcolor=darkblue}
\usepackage[utf8]{inputenc}
\usepackage[T1]{fontenc}
\usepackage[small]{caption}
\usepackage{graphicx}
\usepackage{amsmath}
\usepackage{amsthm}
\usepackage{booktabs}
\usepackage{algorithm}
\usepackage{algorithmic}
\usepackage{enumitem}

\usepackage[normalem]{ulem}

\newcommand{\citet}[1]{\citeauthor{#1}\ \shortcite{#1}}
\newcommand{\citep}{\cite}
\newcommand{\citealp}[1]{\citeauthor{#1}, \citeyear{#1}}
\def\parcite#1{\citep{#1}} 

\newcommand{\Sref}[1]{Section~\ref{#1}}

\title{THEaiTRE: Artificial Intelligence to Write a Theatre Play}

\def\ufal{$^\mu$}
\def\sd{$^\sigma$}
\def\damu{$^\delta$}

\author{
Rudolf Rosa\ufal\and
Ondřej Dušek\ufal\and
Tom Kocmi\ufal\and
David Mareček\ufal\and
Tomáš Musil\ufal\and\\
Patrícia Schmidtová\ufal\and
Dominik Jurko\ufal\and
Ondřej Bojar\ufal\and
Daniel Hrbek\sd\damu\and\\
David Košťák\sd\and
Martina Kinská\sd\and
Josef Doležal\damu\And
Klára Vosecká\damu
\affiliations
\ufal Charles University, Faculty of Mathematics and Physics, Institute of Formal and Applied Linguistics\\
\sd The Švanda Theatre in Smíchov, Prague \\
\damu The Academy of Performing Arts in Prague, Theatre Faculty (DAMU)\vspace{1mm}
\emails
uru@ufal.mff.cuni.cz,
hrbek@svandovodivadlo.cz
}

\begin{document}
  
\maketitle
\thispagestyle{plain}
\pagestyle{plain}

\begin{abstract}
  We present THEaiTRE, a starting research project aimed at automatic generation of theatre play scripts.
  This paper reviews related work and drafts an approach we intend to follow.
  We plan to adopt generative neural language models and hierarchical generation approaches, supported by summarization and machine translation methods, and complemented with a human-in-the-loop approach.
\end{abstract}



\section{Introduction}

We introduce the THEaiTRE project,\footnote{\url{https://www.theaitre.com/}} which aims to produce and stage the first computer-generated theatre play. 
This play will be presented on the occasion of the 100th anniversary of Karel Čapek's play \emph{R.U.R.} \citep{rur}, for which the word “robot” was invented by Čapek.

The project, currently in its early stages, is at the intersection of artificial intelligence research and theatre studies.
The core of our approach is to use state-of-the-art deep neural models trained and fine-tuned on theatre play data.
However, our team includes both experts on natural language processing and theatre experts, and our solution will be based on research and exchange of experience from both fields.

In this paper, we first review related previous works (\Sref{sec:related}) and data resources available to us (\Sref{sec:data}).
We then draft the approaches we are following and intending to follow in the project (\Sref{sec:approach}) and present the project timeline (\Sref{sec:plan}).




\section{Related Work}
\label{sec:related}

\subsection{Narrative Natural Language Generation}
\label{sec:related-nlg}

While we are not aware of any generation systems specifically aimed at theatre play generation, research in story/narrative generation has been quite active in the past years, with computer-aided systems allowing various degrees of automation and different abilities in learning from data \cite{kybartas_survey_2017,riedl_computational_2018}.
Since recurrent neural networks (RNN) were applied for text generation \cite{bahdanau_neural_2015,sutskever_sequence_2014}, research in story generation has mostly focused on fully data-driven, fully automated approaches.
As plain RNNs were found unsuitable for producing longer, coherent texts \cite{wiseman_challenges_2017}, multiple improvements have been proposed.

The first line of work focuses on providing a higher-level semantic representation to the networks and conditioning the generation on it. \citet{martin_event_2018} and \citeauthor{ammanabrolu_guided_2019}~[\citeyear{ammanabrolu_guided_2019}; \citeyear{ammanabrolu_story_2020}]  use an event-based representation, where an event roughly represents a clause (predicate, subject, direct and indirect object). The model generates the story at the event level and subsequently realizes the individual events to surface sentences.
\citet{tu_generating_2019} take a similar approach, using frame semantics and also conditioning sentence generation on other information, such as sentiment.

Other works focus on explicit entity modelling across the generated story, e.g., \citet{clark_neural_2018}. Here, each entity has its own distributed representation (embedding), which is updated on each mention of the entity in the story.

Multiple authors attempt to increase long-term coherence by hierarchical story generation.
\citet{fan_hierarchical_2018} generate first a short prompt/tagline, then use it to condition the full story generation.
\citet{yao_plan-and-write:_2019} take a similar approach, using a ``storyline'' -- a list of entities and items to be introduced in the story in the given order.
\citet{fan_strategies_2019} then combine the hierarchical generation with explicit entity modelling. Their system generates outputs using anonymized but tracked entities, which are subsequently lexicalized in the context of the story by generating referring expressions.

Several works experiment with altering the base RNN architecture:
\citet{wang_t-cvae:_2019} use a modified Transformer architecture \cite{vaswani_attention_2017}, which is trained as a conditional variational autoencoder.
\citet{tambwekar_controllable_2019} utilize reinforcement learning with automatically induced rewards to train their event-based model.
\citeauthor{ammanabrolu_guided_2019}~[\citeyear{ammanabrolu_guided_2019}; \citeyear{ammanabrolu_story_2020}] extend this work by experimenting with various sentence realization techniques, including retrieval from database and post-editing.

Latest works use massive pretrained language models based on the Transformer architecture, such as GPT-2 \cite{radford_language_2019}, for generation.
\citet{see_massively_2019} use GPT-2 directly and show that it is superior to plain RNNs.
\citet{mao_improving_2019} apply GPT-2 fine-tuned for both story generation and common-sense reasoning to improve coherence.

While research in this area has progressed considerably, most experiments have been performed on rather short and simple stories, such as the ROCStories corpus \cite{mostafazadeh_story_2016}. Many works focus on limited tasks, such as single-sentence continuation generation \cite{tu_generating_2019}.
The state-of-the-art results still cannot match human performance, producing repetitive and dull outputs \cite{see_massively_2019}.

\subsection{Dramatic Analysis}


For our needs, we are mostly interested in classifications and abstractions over theatre play scripts or their parts.
In the field of theatre studies, there is a vast amount of research on the structure and interpretation of theatre plays.
Unfortunately, the results of such research are not made available in forms and formats that would easily allow us to use these as data and annotations in machine learning approaches.

The Thirty-Six Dramatic Situations by \citet{polti1921thirty}\footnote{\url{https://en.wikipedia.org/wiki/The_Thirty-Six_Dramatic_Situations}} is a classic work, in which the author presented a supposedly ultimate list of all categories of possible dramatic situations that can occur in a theatre play (e.g.\ ``adultery'' or ``conflict with a god''), further subclassified into 323 situational possibilities.

Although not directly related to theatre plays, the work of \cite{propp1968morphology} is also essential.
Propp analyzed Russian folk tales and identified 31 \textit{functions}, similar to Polti's situations but somewhat more down-to-earth (e.g.\ ``villainy'' or ``wedding''), as well as 7 abstract character types (e.g.\ ``villain'' or ``hero'') and other abstractions.

Polti's and Propp's categorizations are sometimes used in analyzing and generating narratives, although typically not in drama.
The work closest to our focus is probably that of \cite{gervas2016annotating}, who devised an ontology of abstractions for annotating musical theatre scripts, based on both of the mentioned works, as well as on more recent plot categorization studies \citep{booker2004seven,tobias201120}.

There are also works producing drama analyses in the form of networks, capturing various relations between the characters in the play \citep{Moretti2014OperationalizingOT,dracor,gerdracor}.

\subsection{Computer-Generated Art}

There already is a range of partially or fully artificially generated works of art, some of which we list here.
While this demonstrates the technical possibility of such an approach, the mixed reception of the outcomes shows that the employed technologies are clearly not on par with humans (yet?).


\begin{description}

\item[Sunspring] is a short sci-fi movie with an LSTM-generated and human-post-edited script \citep{sunspring}.

\item[Beyond the Fence] is a musical based on suggestions from several automated tools, heavily human-post-edited \citep{colton2016beyond}.

\item[Poezie umělého světa] is a human-picked collection of computer generated poems \citep{materna2016poezie}.

\item[Lifestyle of the Richard and Family] is a theatre play written with the help of a next word suggestion tool \citep{helper2018}.

\item[From the Future World] is a musical composition generated in the style of Antonín Dvořák \citep{dvorak}


\end{description}



\section{Data Resources}
\label{sec:data}

It is necessary to collect as much data as possible to train or at least to fine-tune our system for generating theatre plays, ideally well-formatted theatre scripts or movie scripts. We also collect synopses -- brief summaries of the major points of the plays, which could be used for generating the play hierarchically (see \Sref{sec:hierarchical}).




We focus on the English and Czech languages and, surprisingly, we have more and better data available for Czech. So far, we collected more than 600 plays and 1,500 synopses in Czech and about 150 plays and 100 synopses in English.
However, the quality of the data is not ideal in many cases. The majority of the data is in plain-text or HTML/XML format, but it is often hard to automatically
distinguish character names, lines, scenic notes and scene settings.
A lot of normalization will be required to process many different input formats.
Summaries and plays will need to be paired together since they may come from different sources.

Our plan is to convert as many theatre plays as possible into one common JSON format properly structured into acts, scenes, and dialogues.
We may release a part of our play corpus in this format where copyright restrictions allow us to do so.



\section{Planned Approach}
\label{sec:approach}


We intend to base our approach in generative neural models (\Sref{sec:lm-generation}) applied in a hierarchical setup (\Sref{sec:hierarchical}).
This setup will be complemented by synthesizing training data using summarization techniques (\Sref{sec:summarization}), automated machine translation to allow for working with inputs and outputs in both English and Czech (\Sref{sec:mt}), and a human-in-the-loop style cooperation of the automated system with theatre professionals (\Sref{sec:hitl}).

\subsection{Applying Neural Language Models}
\label{sec:lm-generation}

Large neural \emph{language models} (LMs), such as GPT-2 [\citealp{radford_language_2019}; see Section~\ref{sec:related-nlg}], are able to generate believable texts in certain domains (e.g.\ news articles).
This is not the case for the domain of theatre plays. The original GPT-2 must have had a number of plays (or movie scripts) in the training data, which is evident when it is presented with a suitable starting prompt. It can produce a text that follows the formal structure and has some level of content coherence. However, the basic attributes of a dramatic situation are missing: there is no plot, and the scene is not moving towards a conclusion. Other problems include having new characters appear randomly in the middle of the scene or falling into a state of repeating the same sentence forever.


Our basic workflow would be to seed an LM with a prompt which is the beginning of a dramatic situation. The LM would generate the rest of the whole dialogue.
We plan to finetune the LM to theatre plays to see how far this approach can go.
Then we plan to restrict the generation by enforcing that only certain predetermined characters speak, possibly in a pregenerated order. This can be achieved by stopping the generation at the end of a character's line, adding the name of the next desired character and then resuming the generation process.



To make the characters more internally consistent and different from each other at the same time, we plan to devise individual LMs specialized to specific character types, based on a clustering of the characters across plays.
The part of each character would then be generated by a different LM; i.e., the script would consist of several LMs “talking” to each other.



\subsection{Hierarchical Generation}
\label{sec:hierarchical}

We also plan to extend our experiments with hierarchical generation from large pretrained LMs.
We will use an approach similar to \citet{fan_hierarchical_2018} and \citet{yao_plan-and-write:_2019} (see \Sref{sec:related-nlg}): starting with generating a title or a prompt for the story, then generating a textual synopsis. The generation of the play from the synopsis will follow as a novel step, not present in previous works.
We are considering multiple options of what to choose as the synopsis representation: the play background/setting from play databases, more detailed synopses from fan websites, or scenic remarks extracted from texts of plays themselves. Ultimately, the choice will be made based on data availability.
The synopsis generation step should also include generating the list of characters, which will be adhered to in the final text generation step.

The final step will use a similar approach as the base LM generation (see \Sref{sec:lm-generation}).
We also plan on using explicit embeddings for individual characters in the play and using explicit entity tracking/coreference \citep{clark_neural_2018,fan_strategies_2019}. Since the available automatic coreference tools [e.g., \citealp{clark_deep_2016}; \citealp{lee_end--end_2017}] are typically not trained for processing dialogic texts, they may require adaptation.

\subsection{Synthesizing Data with Summarization Techniques}
\label{sec:summarization}

The hierarchical generation approach relies on data that contain information of various granularity, as described in \Sref{sec:hierarchical}. However, most of the available data contain only the title and the script of the play, missing other invaluable information. In our project, we intend to synthesize the missing data; synthetic data are frequently used in various tasks, such as machine translation \parcite{bojar_improving_2011,sennrich2016backtranslation}.

We can generate synthetic data with the use of the classical task of text summarization; abstractive summarization in particular \parcite{rush-etal-2015-neural}. The main idea is to take a long document and summarize it into a few sentences, then take these synthetic data and use them for training the generative models in the hierarchical approach.
With various summarizing models, we can first abstract the whole script of a theatre play into a detailed synopsis, then the detailed synopsis into a short plot synopsis, and eventually the short synopsis into the play title. With these summarizing models, we can fill the gaps in our datasets, so that the hierarchical generation models can be trained on all theatre scripts available to us, even if they lack some or all higher-level summaries.

We plan to train the Transformer model \parcite{vaswani_attention_2017} for the summarization tasks.
%
As we expect the amount of available training play-summary pairs to be scarce, we will pretrain our models on other summarization tasks, such as news abstract generation for which plenty of parallel data is available \parcite{STRAKA18sumeczech}, followed by fine-tuning the pretrained models on our in-domain theatre data.


\subsection{Machine Translation}
\label{sec:mt}


We plan on using machine translation (MT) for two purposes:
(1) Since we have limited amounts of training data scattered across both English and Czech, we need the generation to take advantage of data in both languages. Therefore, we plan to generate new training data by translating either Czech texts to English or vice versa. (2) We would like the same resulting generated play to be available instantly in both languages. Therefore, we plan to generate it in one of the languages and use MT to bring the result over to the other language. 

For both applications, we are going to use our in-house state-of-the-art Czech-English model \parcite{popel2018wmt}.
However, theatre play scripts are a specific domain of data for which our MT models were not specifically trained. To tackle this problem, we will finetune \parcite{barone2017regularization} the general MT models on theatre parallel data.

\subsection{Human in the Loop}
\label{sec:hitl}



To ensure a satisfactory result, we intend to complement the automated generation with interventions from theatre professionals, using a \textit{human-in-the-loop} approach.
We envision a symbiotic cooperation of human and machine where the machine does most of the work but asks the human for input when necessary.
The exact division of labour between the human and artificial components will emerge through time; two likely places for human interaction are for defining inputs and for picking or reranking the outputs, but other variants are also possible.

Moreover, only the script of the play will be autogenerated. The subsequent realization and performance of the play will be in the hands of theatre professionals.

\section{Conclusion and Future Work}
\label{sec:plan}

The project is currently at its official start, with some preliminary work done.
The approaches described in this paper will be actuated throughout the course of the year 2020.
The first automatically generated THEaiTRE play will be premiered in January 2021, at the occasion of the 100th anniversary of the premiere of the play \textit{R.U.R.} \citep{rur}.
A premiere of a second play, generated from an improved version of our system, is planned for January 2022.

The project can be tracked at its website:
\begin{center}
\url{https://theaitre.com}    
\end{center}





\section*{Acknowledgments}

The THEaiTRE project is supported by the Technology Agency of the Czech Republic grant TL03000348.






\bibliographystyle{named}
\bibliography{ijcai20}

\end{document}